\documentclass[10pt,twocolumn,letterpaper]{article}

\usepackage[review]{cvpr}      % To produce the REVIEW version
\usepackage{footmisc}
\usepackage[dvipsnames]{xcolor}

\definecolor{cvprblue}{rgb}{0.21,0.49,0.74}
\usepackage[pagebackref,breaklinks,colorlinks,citecolor=cvprblue]{hyperref}
\usepackage{multirow}
\title{
Active Generation Network of Human Skeleton for Action Recognition
}

\author{Long Liu \footnote{Corresponding author}\\
{\tt\small liulong@xaut.edu.cn}
\and
Xin Wang\\
{\tt\small wangxin2168@163.com}
\and
Fangming Li\\
{\tt\small 2220321220@stu.xaut.edu.cn}
\and
Jiayu Chen\\
{\tt\small cjy21123@163.com}
}

\begin{document}
\maketitle
\begin{abstract}

Data generation is a data augmentation technique for enhancing the generalization ability for skeleton-based human action recognition. Most existing data generation methods face challenges to ensure the temporal consistency of the dynamic information for action. In addition, the data generated by these methods lack diversity when only a few training samples are available. To solve those problems, We propose a novel active generative network (AGN), which can adaptively learn various action categories by motion style transfer to generate new actions when the data for a particular action is only a single sample or few samples. The AGN consists of an action generation network and an uncertainty metric network. The former, with ST-GCN as the Backbone, can implicitly learn the morphological features of the target action while preserving the category features of the source action. The latter guides generating actions.  Specifically, an action recognition model generates prediction vectors for each action, which is then scored using an uncertainty metric. Finally, UMN provides the uncertainty sampling basis for the generated actions.

\end{abstract}    
\section{Introduction}
\label{sec:intro}

Human action recognition (HAR) is one of the hotspots in computer vision, widely applied to intelligent surveillance, human-computer interaction, and virtual reality \cite{aggarwal2014human,escalera2017challenges,han2017space,wang2018rgb}.
The main methods are RGB-based, RGBD-based, and Skeleton-based \cite{ji20123d,feichtenhofer2016convolutional,feichtenhofer2019slowfast,carreira2017quo,yang2019asymmetric,li2019actional}. 
In contrast, the Skeleton is a simple structure that is robust to changes in appearance features, complex backgrounds, and occlusion interference in RGB data. Therefore, Skeleton-based HAR is gradually becoming a mainstream method.

In recent years, graph convolutional networks (GCN) \cite{yan2018spatial,shi2019two,bai2022skeleton,cai2021jolo} have rapidly developed to extract spatio-temporal relationships among joints.
Applications of GCNs have achieved outstanding performance in skeleton-based HAR.
These results are largely dependent on the availability of large amounts of data.
However, human action data can only be obtained for a few or even one sample because of privacy, low probability of occurrence, and high cost, e.g., cheating on exams, robbery, and homicide.
These issues limit the quantity of human action data, thereby limiting the generalization ability of HAR. With the continuous development of data generation techniques, the possibility of generating large datasets has emerged. 

The traditional data generation methods include geometric transformations, noise injection, and data interpolation \cite{shorten2019survey,he2019parametric}. The original data limits this method, cannot generate new data, and is easily distorted. 
Subsequently, many deep learning-based data generation methods have been proposed, such as generative adversarial networks (GAN) \cite{choi2018stargan,choi2020stargan,karras2019style,karras2020analyzing,wang2018unregularized}, variational autoencoders (VAE) \cite{kingma2013auto,lopez2018information,lopez2020decision}, flow models \cite{rezende2015variational,kingma2018glow}, and diffusion models \cite{ho2020denoising,liu2022compositional,zhang2023inversion}. 
These methods are effective only for generating static data such as images and text \cite{liu2019wasserstein, radford2015unsupervised, mao2017least}. However, for dynamic data such as human actions, the generated actions may be unnatural and discontinuous between frames and lack temporal consistency. 
In addition, when only a few training samples are available, it leads to a poor generalization of the generated model, and the generated data lacks diversity.
Few-shot generation approaches \cite{antoniou2017data,hong2022deltagan,clouatre2019figr,hong2020matchinggan,hong2020f2gan,finn2017model,nichol2018reptile} aim to generate a large amount of natural and diverse data for a few new categories, partly solving this problem. 
However, the method mainly uses a small number of samples for data generation and does not utilize the rich information of many base categories. Many base category action for human actions have more complete data distribution information. Incorporating this feature information into the generation process can provide richer features for new actions.

To address the above, motion style transfer provides an excellent solution. Motion style transfer \cite{jang2022motion, aberman2020unpaired, ParkSoomin2021Diverse, 2016A} aims to extract the target style from a action example and transfer it to another action with the desired content. The problem of temporal consistency is improved by the adaptive instance normalization (AdaIN) \cite{huang2017arbitrary,saito2020coco} aligning the two action features rather than simply fusing them.
The recent work by JANG et al. \cite{jang2022motion} proposes a novel motion style transfer network. The network consists of multi-layer ST-GCNs that can achieve arbitrary motion transfer without style labeling. Our work follows the same approach.
Motion Puzzle divides the human skeleton into five parts, allowing flexible control over the migration of specified parts during generation. This approach is effective for single-action generation tasks.
However, it is usually time-consuming to control parts for generation when generating many actions. In addition, Motion Puzzle's target motion encoder is connected to the decoder at multiple scales, which may constrain the diversity of action.
Although motion style transfer is a generative network, it is not designed to solve the problem of data sparsity but is purely a one-to-one feature transfer. The quality of generation is only partially guaranteed when generating lots of data. 
To this end, we select and utilize the most informative samples by incorporating active learning \cite{du2021contrastive, sinha2019variational, zhang2020state} to guide human action generation.

In this paper, a novel action generation network called Active Generation Network (AGN) is proposed for skeleton data generation. Our method adaptively learns various action categories by motion style transfer. With only a few or even a single sample, AGN can generate many new actions without assigning body parts. A unique advantage is incorporating active learning into the generation process. For a large number of action samples generated, the most valuable samples are implicitly selected using an uncertainty metric in active learning to ensure the quality of the generation. To the best of our knowledge, our work is the first that guides the generation of human actions using active learning.

The AGN consists of a action generation network (AcGN) and an uncertainty metric network (UMN). The MGN consists of two encoders and a decoder. The encoders extract action features by the graph convolution layer and instance normalization layer, and then the decoder synthesizes new actions. The MGN can implicitly learn the skeletal morphology of the target actions without stylizing any actions while preserving the categories of the original actions. Inspired by active learning, we developed UMN to guide the MGN. 
Firstly, we train ST-GCN using a few or a single sample to generate prediction vectors for new actions. Then, a score is obtained from the uncertainty metric, based on which samples are selected and added to the train set to train the ST-GCN again. 
This process is repeated until the data meets the requirements. 

The contributions of our work can be summarized as follows:

\begin{itemize}
    \item We propose a generative network called MGN for skeletal data to generate high-quality human action data with only a few or a single sample.
    \item We propose AGN for Human Skeletons for HAR. Introducing active learning into the generation process implicitly selects the most valuable samples using an uncertainty metric to ensure the generation quality.
    \item FMD and Accuracy are used to evaluate the results on the NTU-RGB+D dataset. The results show that our method is competitive with other methods. The method requires only 10\% of the original data for the same accuracy.
\end{itemize}

\section{Related work}

\textbf{Generative Adversarial Networks. }
Generative Adversarial Networks (GAN) is a generative models which is trained by adversarial learning. 
In the early days, unconditional GAN \cite{karras2019style,karras2020analyzing} recovered images from random noise. Developed to the present, conditional GAN \cite{pan2023drag} utilizes text and images for guidance to generate images. GAN has performed strongly on tasks of generating static data such as image generation \cite{karras2019style, karras2020analyzing}, image editing \cite{vinker2021image, patashnik2021styleclip}, and image translation \cite{shao2021spatchgan}.
The generation of dynamic data, such as videos and action sequences, has also been studied. Carl et al. \cite{vondrick2016generating} proposed a video generation network with a spatial-temporal two-stream convolutional architecture based on DCGAN \cite{radford2015unsupervised}. This work is the first application of GAN to video generation. 
TGAN \cite{saito2017temporal} followed, which first generates a set of latent vectors from noise vectors, then generates pictures and synthesizes videos separately.
RNN-GAN \cite{mogren2016c} is based on the temporal modeling capability of RNNs to predict video from a single frame. It has a more robust motion prediction capability compared to the work of Carl et al. However, these impressive results are mainly attributed to the support of many training samples. With limited data, GANs are prone to overfitting, leading to a lack of diversity in the generated data. 

% \textbf{Few-shot Generation. }

\textbf{Motion Style Transfer. }
Image Style Transfer \cite{gatys2016image, saito2020coco} combines style and content features from two images to form a new image. Motion Style Transfer refers to Image Style Transfer to form a new action by transferring one action's style features to another that contains only content features. Early motion style transfer was done by manually defining style features and inferring them through machine learning \cite{xia2015realtime, yumer2016spectral}. This method is effective only for the actions in the training data with limited scope of usefulness. 
Deep learning-based methods have greatly improved the quality and application of motion style transfer. Both Holden et al. \cite{2016A} and Du et al. \cite{du2019stylistic} applied the Gram matrix method to convey motion styles through the distribution of actions in the hidden space. These methods are time-consuming and have limited the quality of action generation for relatively significant motion differences.
Recently, Aberman et al. \cite{aberman2020unpaired} proposed a motion transfer network that combines GAN and AdaIN. 
The method can learn from unpaired data with different styles to migrate model unseen actions. 
Park et al. \cite{ParkSoomin2021Diverse} used a spatio-temporal graph convolutional network to model actions. The method adds random noise in the decoder to enhance action diversity. 
Jang et al. \cite{jang2022motion} proposed a novel motion style transfer network called Motion Puzzle. 
Motion Puzzle divides the human skeleton into five parts, allowing flexible control over the migration of specified parts during generation. This approach is effective for single-action generation tasks.
However, it is usually time-consuming to control parts for generation when generating many actions. In addition, Motion Puzzle's target motion encoder is connected to the decoder at multiple scales, which may constrain the diversity of action.

\textbf{Active Learning. }
Existing active learning methods are categorized into pool-based and synthetic methods \cite{gal2016dropout,beluch2018power,gorriz2017cost,yang2017suggestive,nguyen2004active}. Pool-based methods use different sampling strategies to determine how to select the most informative samples, with uncertainty sampling methods being the most common.
Ebrahimi et al. \cite{ebrahimi2019uncertainty} used a Bayesian neural network for uncertainty evaluation. Gal \cite{gal2016dropout} and Gharamani \cite{gal2017deep} also showed the relationship between uncertainty and dropout to estimate uncertainty in neural network prediction.
Pool-based methods select samples conditional on a large amount of unlabeled data. In the case of scarcity of data, synthetic methods are more suitable than pool-based methods. Synthetic methods use a generative model to generate samples, then sample based on the uncertainty of the model.
The work of Zhu et al.  \cite{zhu2017generative}, Mahapatra et al.  \cite{mahapatra2018efficient}, and Mayer et al. \cite{mayer2020adversarial} uses GAN to generate a sample and then query using the uncertainty principle. Our work uses this same strategy to guide human action generation using the amount of sample information.

\begin{figure*}[htpb]
  \centering
   \includegraphics[width=0.8\linewidth]{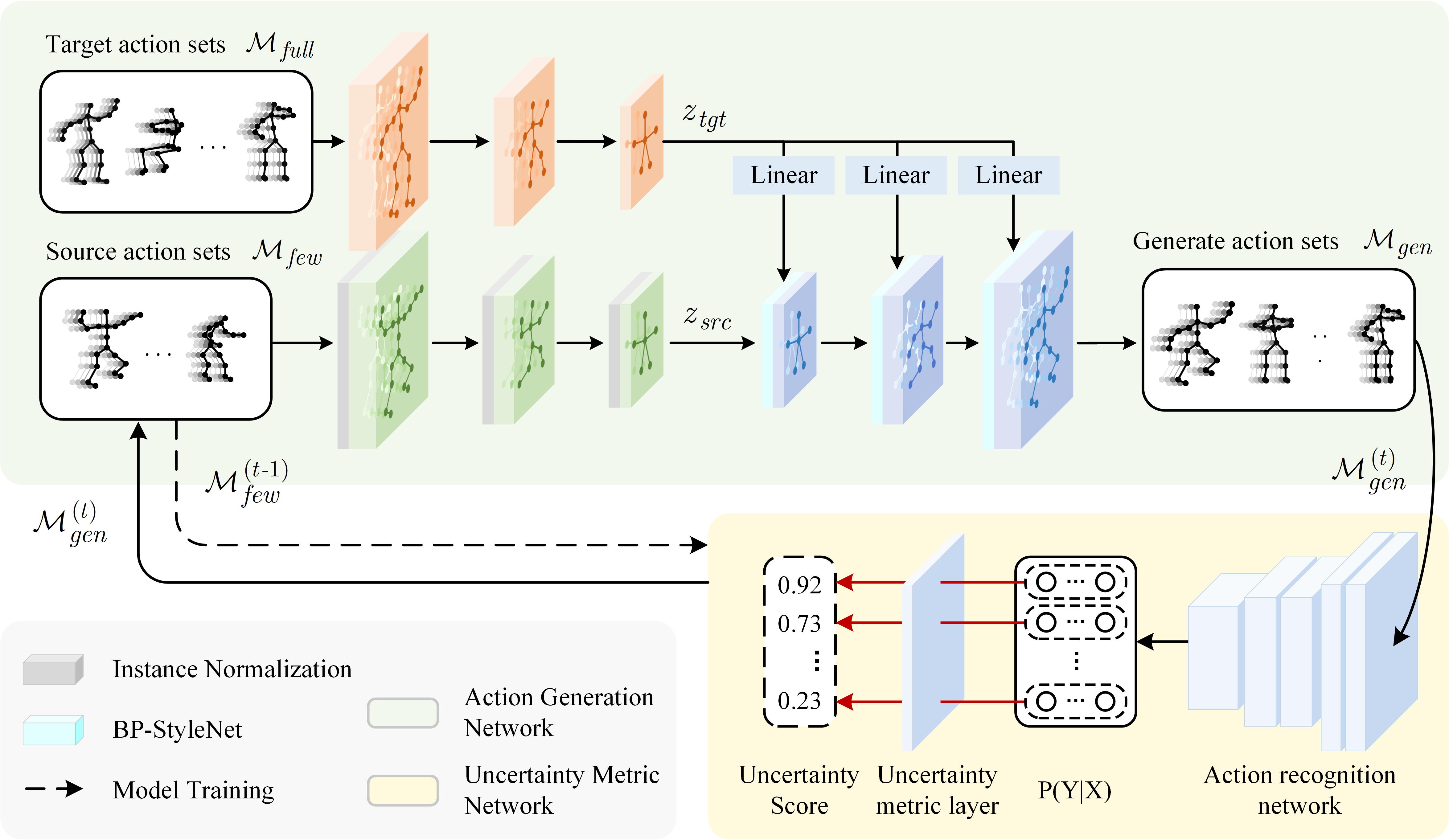}

   \caption{The overall network architecture of our AGN framework. }
   \label{fig:1}
\end{figure*}

\section{Method}

\subsection{Overview}

Figure \ref{fig:1} shows the overall architecture of the AGN framework, consisting of MGN and UMN. 
The MGN generates new actions, and the UMN evaluates the generated actions and inversely guides the generation of MGN. 
We construct a human action set $\mathcal{M}=\mathcal{M}_{train}\cup \mathcal{M}_{unseen}$ using 3D skeletal data from NTU-RGB+D 60  \cite{shahroudy2016ntu}, where $\mathcal{M}_{train}\cap \mathcal{M}_{unseen}=\varnothing$. 
In addition, we define different action subsets. 
The AGN's input is a complete target action set $\mathcal{M}_{full}$ and a one-shot or few-shot source action set $\mathcal{M}_{few}$. The final output is a complete action set $\mathcal{M}_{gen}$ of equal size $\mathcal{M}_{full}$. The action is denoted as $\textbf{M}_{src}\in\mathcal{M}_{few}, \textbf{M}_{tgt}\in\mathcal{M}_{full}$, and $\textbf{M}_{gen}\in\mathcal{M}_{gen}$. 

The MGN uses spatio-temporal graph convolutional layers as the basis to construct the encoder and decoder, connecting the high and low dimensional feature layers of the human skeletal graph structure by graph upsampling and downsampling \cite{yan2019convolutional, jang2022motion}.
The encoder extracts features for the skeletal morphology and the category information of the action, respectively. The decoder outputs the new action by fusing the multi-scale spatio-temporal features of the source and target actions via the BP-StyleNet Layer \cite{jang2022motion}.
The UMN guides the MGN to generate high-quality action. 
Specifically, the UMN obtains the a posteriori probability of each action via a recognition network. Then, the uncertainty score is obtained via an uncertainty metrics layer to provide a basis for sample selection.

\subsection{Action Generation Network}

\textbf{Graph Upsampling and Downsampling. }
A practical method for extracting features in image generation is progressive upsampling and downsampling. The upsampling gradually improves the image resolution and increases the local details, and the downsampling can aggregate the image features and reduce the noise. 
The upsampling and downsampling are generally implemented through Unpooling and Pooling. It proved effective in graph convolutional neural networks for skeletal data\cite{jang2022motion, yan2019convolutional, ParkSoomin2021Diverse}. 
Following this idea, we incorporate the upsampling and downsampling methods combined with information entropy into the spatial-temporal graph to extract local and global features. 

\textbf{Action Encoder. }
We developed the action encoder similar to VGG16 \cite{simonyan2014very} to map human actions to latent space using graph upsampling and spatio-temporal graph convolutional layers. Given an action $\textbf{M}_s$, the encoding process is written as: 

\begin{equation}\label{e2}
z_s=E_s(\textbf{M}_s),
\end{equation}

\noindent where $s\in \{src, tgt\}$.
The action encoder consists of a source encoder and a target encoder. 
Each encoder $E_s$ is a concatenation of multi-level encoding blocks $E_s^i$ to gradually extract the latent feature $z_s^i=E_s^i(z_s^{i-1})$, where $i\in \{1,2,3\}$, and $z_s^i$ is the action feature obtained after each encoding block. 

The source encoding block $E^i_{src}$ consists of instance normalization layer (IN), ST-GCN, and graph downsampling (GD) to extract source action features gradually. 
In the target encoder, we wish to preserve the morphological features of the action to combine with the category features of the source action to form a unique new action. Therefore, it consists only of ST-GCN and GD. 

\textbf{Feature Decoder. }
The feature decoder fuses the category features $z_{src}$ of the source action with the morphological features $z_{tgt}$ of the target action to synthesise a new action $\textbf{M}_{gen}$. The dncoding process is written as: 

\begin{equation}\label{e5}
\textbf{M}_{gen}=D(z_{src}, z_{tgt}).
\end{equation}

\noindent The decoder is similar in structure to the encoder and consists of three decoding blocks and three linear layers. 
The three decoding blocks recover the action sequences step-by-step by fusing $z_{tgt}$ and $\hat{z}^0_{dec}(=z_{src})$, defined as:

\begin{equation}\label{e6}
\hat{z}^i_{dec}=D^i(\hat{z}^{i-1}_{dec}, L^i(z_{tgt})),
\end{equation}

\noindent where $i\in \{1,2,3\}$, $\hat{z}_{dec}^i$ is the action feature output from each decoding block, and $L^i$ is the linear layer mapping the deep feature $z_{tgt}$ to the same feature map size as the output of each decoding blocks.

We use Body Part Adaptive Instance Normalisation (BP-AdaIN) and Body Part Attention Network (BP-ATN) in the decoding block from Motion Puzzle \cite{jang2022motion}, where BP-AdaIN applies AdaIN \cite{gatys2016image, saito2020coco} according to the body parts, extending the network's degree of freedom, and more flexible fusion of the features of each part of the target and the source action. 
The BP-ATN constructs the feature attention mapping of the target and source actions. 
BP-AdaIN and BP work together to extract local and global features of the target action.

\subsection{Uncertainty Metric Network}

\textbf{Action Recognition Network}. 
The prediction vectors are obtained from the action recognition model, thereby calculating the uncertainty score. 
Our task is oriented towards data generation for action recognition, and thus, ST-GCN is adopted as the task model. 
At the first iteration, $\mathcal{M}^{(1)}_{gen}=MGN(\mathcal{M}^{(0)}_{few},\mathcal{M}_{full})$ is obtained by inputting $\mathcal{M}^{(0)}_{few}$ into the MGN. 
Meanwhile, the task model is trained using $\mathcal{M}^{(0)}_{few}$. Finally, prediction vectors are generated for each action. 
Starting from the second iteration, $\mathcal{M}^{(t)}_{few}=\mathcal{M}^{(t-1)}_{few}\cup\mathcal{M}^{(t)}_{gen}, t\in[1,iter]$. 
Given that the number of categories is $L$, then $Y = p(\mathcal{M}^{(t)}_{gen}|\mathcal{M}^{(t-1)}_{few})\in\mathbb{R}^{K\times L}$,  where $K$ is the number of generating samples, and $p(A|B)$ denotes the prediction vectors produced by the action recognition model trained under dataset $B$ for dataset $A$.

\textbf{Uncertainty Metrics}. 
The uncertainty score of the prediction vector is calculated by the uncertainty metric. The uncertainty score is calculated as follows:

\begin{equation}\label{e7}
S(Y)=I-\frac{Var(Y'[k])}{Var(Y[k])}\times max(Y[k]),
\end{equation}

\noindent where $I$ is a full one-vector of length $K$, then $S(Y)\in\mathbb{R}^K, k\in [1,K]$. The $Var(Y[k])$ can be formulated as:

\begin{equation}\label{e8}
Var(Y[k])=\frac{1}{L}\sum_l (Y[l,k]-\frac{1}{L})^2.
\end{equation}

\noindent The $Var(Y'[k])$ is the minimum variance of the same vector as the maximum value of $Y[k]$, denoted as:

\begin{equation}\label{e9}
\begin{aligned}
&Var(Y’[k])=\\
&\frac{1}{L}((max(Y[k])-\frac{1}{L})^2+(L-1)(\frac{1-max(Y[k])}{L-1}-\frac{1}{L})^2).
\end{aligned}
\end{equation}

\noindent The maximum value in $Y'[k]$ is the same as the maximum value of $Y[k]$, and the other elements are $\frac{1-max(Y[k])}{L-1}$. 
Therefore, $\frac{Var(Y'[k])}{Var(Y[k])}$ represents the degree of concentration of the probability distribution of the predicted vectors. It ensures that each score ranges from 0 to 1 and is negatively correlated with the maximum vector, i.e., a smaller $max(Y[k])$ indicates greater uncertainty.
Finally, the action data is selected based on this score to form the action set $\mathcal{M}_{gen}$.

\subsection{Training}

We train the action generation network end-to-end, given the source action $\textbf{M}_{src}\in\mathcal{M}_{src}$ and the target action set $\textbf{M}_{tgt}\in\mathcal{M}_{tgt}$, and optimize the network with the following loss function. 

\textbf{Reconstruction loss} and \textbf{Cycle consistency loss} are outstanding in motion style transfer \cite{jang2022motion, aberman2020unpaired, ParkSoomin2021Diverse, 2016A}.
Reconstruction loss gives the network the ability to reconstruct a movement. For each action in the action set, the network can reconstruct the action after feature disentanglement and feature fusion. The new action should have both source action category features and target action morphological features. Therefore, the encoders $E_{src}$ and $E_{tgt}$ are used to disentangle the features of $\textbf{M}_{gen}$ and the acquired features are used to compute the cycle consistency loss with the features of the source and target actions, respectively. 

\textbf{Feature triplet loss. }
In order to make the category features of the action more apparent in the latent space, a triplet loss is used during training to make the same category of actions clustered with each other and different categories of actions far away from each other so that the network captures the similarities and differences between action features. 

\begin{equation}\label{e12}
\begin{aligned}
&\mathcal{L}_{trip}=\\
&\mathbb{E}_{\textbf{M}^t_i,\textbf{M}^t_j,\textbf{M}^s_k\sim\mathcal{M}}(||E_{src}(\textbf{M}^t_i)-E_{src}(\textbf{M}^t_j)||-\\
&||E_{src}(\textbf{M}^t_i)-E_{src}(\textbf{M}^s_k)||+\delta),
\end{aligned}
\end{equation}

\noindent where $\textbf{M}^t_i$ and $\textbf{M}^t_j$ represent two motions of the same category,  $\textbf{M}^t_{i,j}$ and $\textbf{M}^s_k$ denote two motions of the different category, so $t\neq s, i\neq j\neq k$. The boundary value $\delta=5$. 

The total objective function of the MGN is thus: 

\begin{equation}\label{e13}
\mathcal{L}_{total}=\lambda_{rec}\mathcal{L}_{rec}+\lambda_{cyc}\mathcal{L}_{cyc}+\lambda_{trip}\mathcal{L}_{trip},
\end{equation}

\noindent where $\lambda_{rec}, \lambda_{cyc},$ and $\lambda_{trip}$ are the hyperparameters of each loss term. 1, 0.5, and 0.5, respectively, in our experiments.

\section{Experiments}

We conducted various experiments to prove the effectiveness of the present method. 
Firstly, we qualitatively measure the results of our method on seen and unseen data, including action visualization and data downscaling visualization. 
Secondly, the generation quality and accuracy of action recognition were quantitatively measured for the six categories of target actions.
Finally, we performed comparison experiments with previous methods and ablation experiments with a special training loss term.
In addition, we train the ST-GCN using generated and real data, respectively, and test the accuracy of the same real data, thus evaluating the degree of approximation between the generated and real data. 

\subsection{Action Generation}

\begin{figure*}[htpb]
  \centering
   \includegraphics[width=0.75\linewidth]{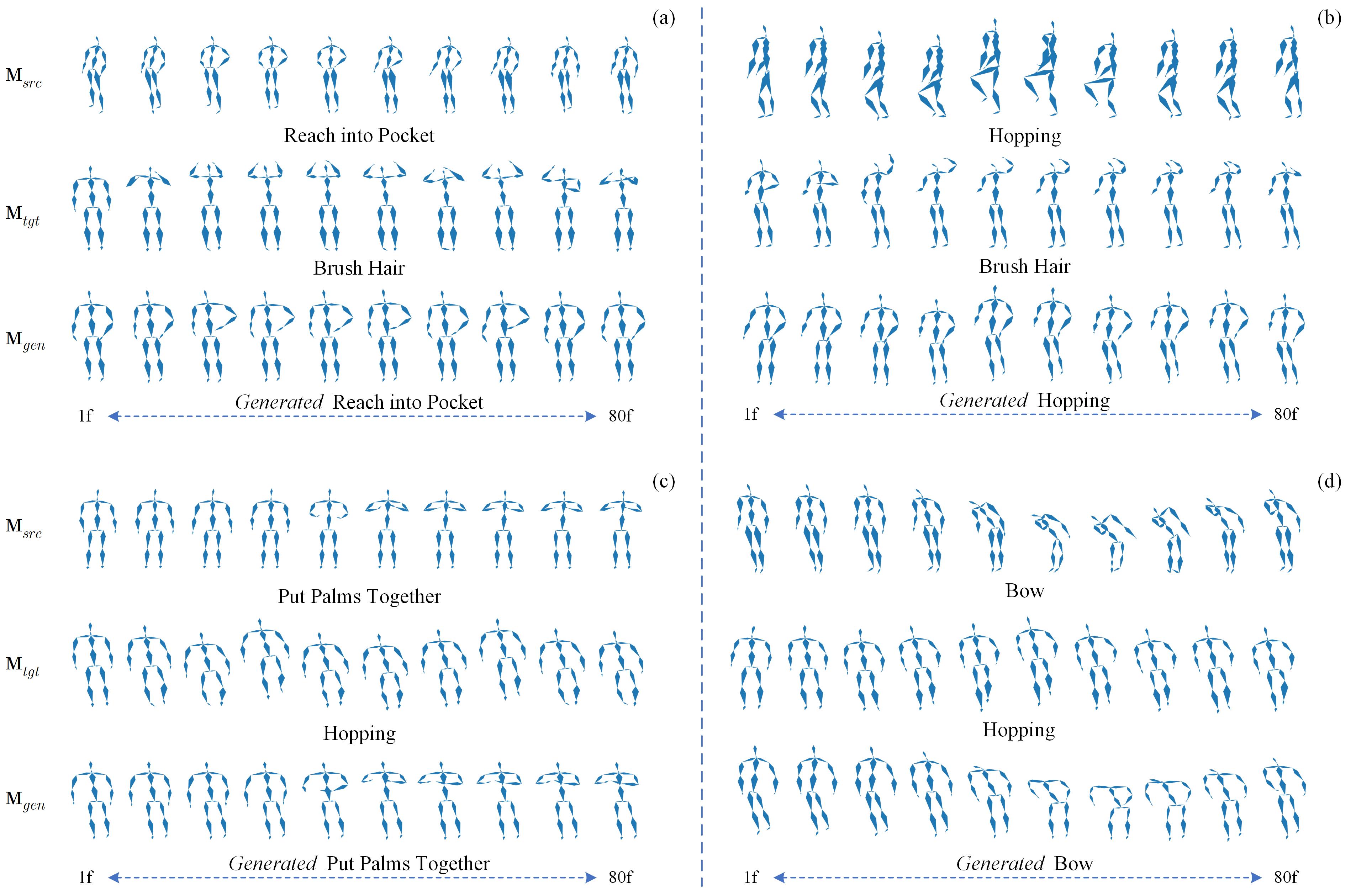}

   \caption{Results generated by MGN on seen actions. (a) “Reach into Pocket”. (b) “Hopping”. (c) “Put Palms Together”. (d) “Bow”.}
   \label{fig:4}
\end{figure*}

\textbf{Qualitative evaluation. }
Figure. \ref{fig:4} shows the generated seen actions. 
(a) and (b) are the “Reach into Pocket” and “Hopping” actions generated with reference to “Brush Hair”, respectively, the former being a hand motion and the latter a whole-body motion. 
(c) and (d) are the “Put Palms Together” and “Bow” actions generated with reference to “Hopping”, respectively. 
All the above actions preserve the source action $\textbf{M}_{src}$ category features and target action $\textbf{M}_{tgt}$ morphological features. 
Compared with Motion Puzzle, this method can generate hand, upper limb, and whole body actions without specifying body parts. 
Meanwhile, the temporal consistency of the actions is guaranteed, e.g., the real and generated “Hopping” are jumping at the same time in (b), and the bending tendency of the generated and real actions are consistent in (d). 

\begin{figure*}[htpb]
  \centering
   \includegraphics[width=\linewidth]{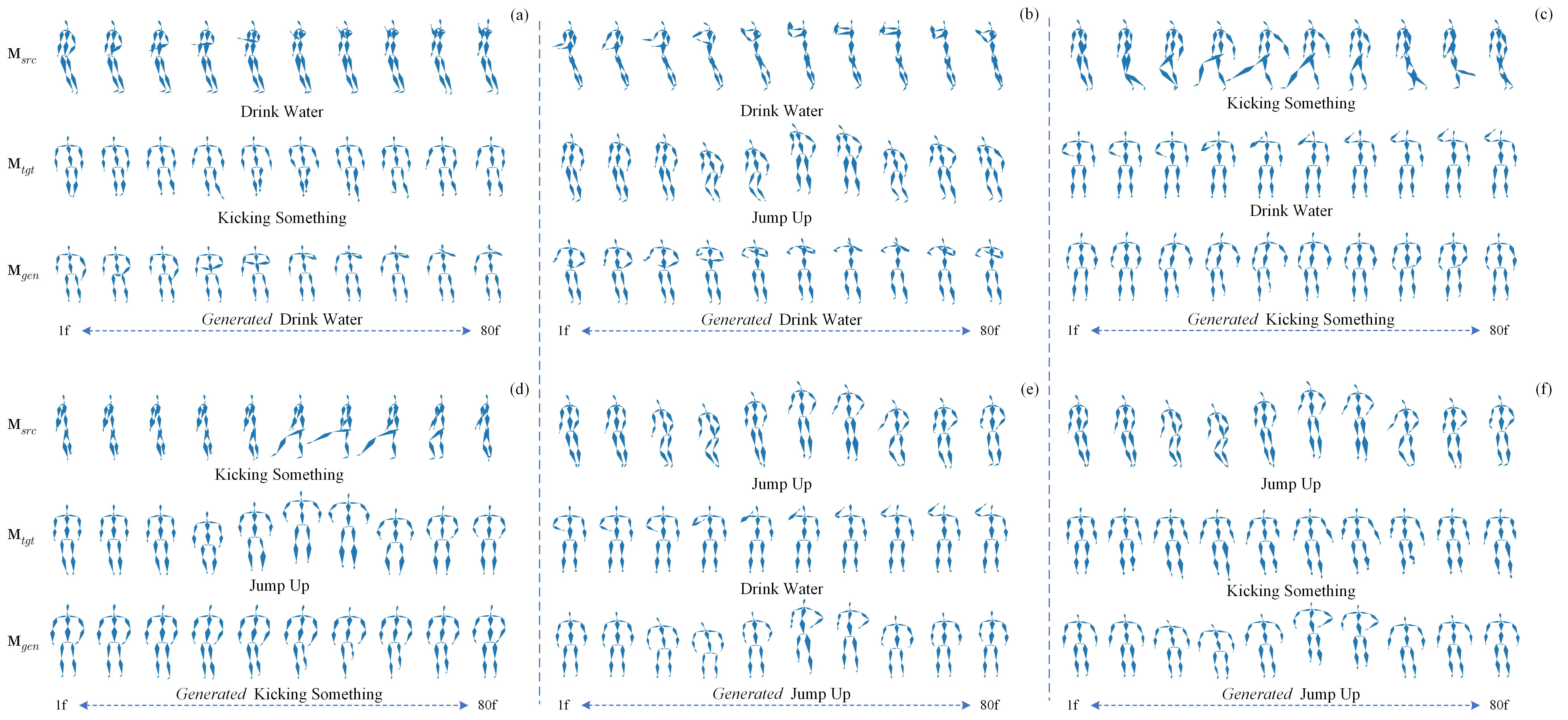}

   \caption{Results generated by MGN on unseen actions. (a) and (b) are  “Drink water” (Unseen). (c) and (d) are “Kicking Something” (Seen). (e) and (f) are “Jump Up” (Unseen).}
   \label{fig:5}
\end{figure*}

Figure. \ref{fig:5} shows the generated unseen motions. 
It contains three cases: only the source action is unseen, only the target action is unseen, and both are unseen to thoroughly verify the transfer effect of unseen actions. 
In (a) and (f), the source action $\textbf{M}_{src}$  (“Drink Water” and “Jump Up”) is unseen, while the target action $\textbf{M}_{tgt}$  (“Kicking Something”) is seen. The generated action $\textbf{M}_{gen}$ can keep the category information of the source action. 
In (c) and (d), $\textbf{M}_{src}$  is seen, and $\textbf{M}_{tgt}$ is unseen. In $\textbf{M}_{gen}$, the morphological features of $\textbf{M}_{tgt}$ are transferred, and the “Kicking Something” action of $\textbf{M}_{src}$ is retained. 
Both source and target actions are unseen in (b) and (e). $\textbf{M}_{gen}$ shows that the model is still able to extract the category information of $\textbf{M}_{src}$ and the morphology information of $\textbf{M}_{tgt}$ to form a new action.
From the generated actions in Figure. \ref{fig:5}, our model can still generate high-quality actions that are unseen for the model.

\begin{figure*}[htpb]
  \centering
   \includegraphics[width=0.8\linewidth]{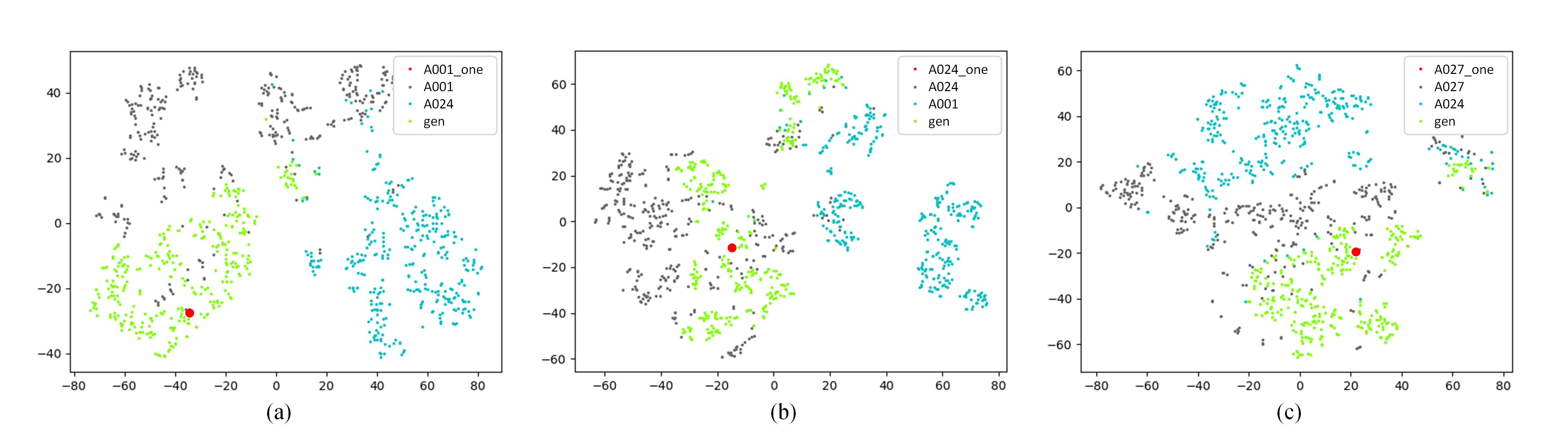}

   \caption{Action data is projected into 2D space using t-SNE, where black is the source action, red is a sample of black, cyan is the target action, and green is some new action generated using a sample of red and some cyan. The green sample and the black sample are very close to each other in space, which indicates that the generated actions conform to some extent to the distribution of the source actions.}
   \label{fig:3}
\end{figure*}

In order to verify the approximation between the generated data and the real data, t-SNE was used to visualize the action data. 
Figure. \ref{fig:3} shows the data distribution after dimensionality reduction using t-SNE. The black and red samples in the figure are the source actions, where red is a random sample in black. Cyan samples are the target actions, while green are the generated samples. The six figures show that the distribution of new actions generated using only one source action is similar to the distribution of source actions. The results show that our generated data can replace the original data. 

\textbf{Quantitative evaluation. }
We quantitatively measured the quality of generation and accuracy of action recognition on seen and unseen data. 
Specifically, we use two metrics: Fréchet Motion Distance (FMD) and Accuracy (Acc). 
We compute FMD and Acc on the action set based on all possible combinations of source and target actions generated by the MGN.

The FMD measures the similarity between the feature vectors of real and generated actions, similar to the Fréchet Inception Distance (FID). 
The action classifier is trained by the ST-GCN method, and feature vectors are obtained after the maximum pooling layer to compute the FMD of generated and real actions. 
A lower FMD means a higher quality of action.

% As described in Sec. \ref{sec:Dataset}, 
We complete the experimental evaluation using the action sets $\mathcal{M}_{full}$, $\mathcal{M}_{seen}$, and $\mathcal{M}_{unseen}$. 
The $\mathcal{M}_{seen}$ contains six categories of seen actions data: “Brush Teeth”, “Pick up”, “Reading”, “Take off a Hat”, “Kicking Something”, and “Sneeze”, for 3996 samples. 
The $\mathcal{M}_{unseen}$ contains six categories of unseen actions data: “Drink Water”, “Throw”, “Sit Down”, “Clapping”, “Jump up”, and “Bow”, for 3992 samples.
Table \ref{tab:1} shows that the Acc of the generated actions all exceeded 90\%. 
The highest of these is 95.39\%, with an average of 92.18\% under the seen actions data. 
The highest of these is 97.20\%, with an average of 94.42\% under the unseen actions data. 
The mean values of FMD in both cases are 2.11 and 2.67, respectively. This shows that our generated actions are high-quality and can be well recognized by the action classifier. 

\begin{table}
  \centering
  % \begin{tabular}{@{}lc@{}}
  \resizebox{1\columnwidth}{!}{
  \begin{tabular}{cccccccc}
    \toprule
    \multirow{2}*{}&\multirow{2}*{Metric}&\multicolumn{6}{c}{Target motion}\\
\cline{3-8}
    &  &  A0&  A1&  A2&  A3&  A4& A5\\
    \midrule
    \multirow{2}*{Seen}& Acc(\%)&  91.10&  90.12&  94.63&  92.29&  95.39& 89.54\\  
    & FMD &  2.19&  2.00&  1.95&  2.02&  2.18& 2.31\\ 
    \multirow{2}*{Unseen}& Acc(\%)&  97.20&  96.45&  89.62&  90.78&  96.77& 95.72\\
    & FMD &  3.21&  2.49&  2.53&  3.11&  2.30& 2.35\\
    \bottomrule
  \end{tabular}}
  \caption{Quantitative evaluation. We calculated FMD and Acc using $\mathcal{M}_{full}$, $\mathcal{M}_{seen}$, and $\mathcal{M}_{unseen}$. $\mathcal{M}_{full}$ contains six categories of action: “Brush Hair”, “Writing”, “Put on a Shoe”, “Take off Glasses”, “Hopping”, and “Shake Head”. For representation simplicity, we numbered the six categories of target action as A0, A1, A2, A3, A4, and A5.}
  \label{tab:1}
\end{table}

\subsection{Action Recognition}
\label{sec:motion recognition}

Generating compelling and high-quality data is significant for action recognition tasks when specific action categories are scarce. 
In order to measure the degree of similarity between generated and real data fully, an action recognition model is trained using generated and real actions, respectively. 

We divided the actions in $\mathcal{M}_{unseen}$ into a training set $\mathcal{M}_{\hat{train}}$ and a test set $\mathcal{M}_{\hat{test}}$. An action recognition model (ST-GCN) is trained using $\mathcal{M}_{\hat{train}}$ and tested on $\mathcal{M}_{\hat{test}}$. As shown in the Table \ref{tab:2}, the top-1 accuracy is 91.80\%.  
We sample one-shot and few-shot (1\%, 5\%, and 10\%) from $\mathcal{M}_{\hat{train}}$ for action generation. 
Subsequently, the generated and sampled actions are concatenated into a new train set to train and test the ST-GCN. 
As shown in Table \ref{tab:2}, the top-1 accuracy is highest at 91.62\% when sampling 10\%, which is only 0.18\% lower than that of $\mathcal{M}_{\hat{train}}$. When sampling 1\% and 5\%, the top-1 accuracy is still high, close to 90\%. 
However, the top-1 accuracy is lower when sampling one, with a maximum of only 62.84\%. 
This result is expected because when sampling a single sample, the generative model is very limited to learning the source data distribution, leading to a large deviation of the generated data distribution from the original complete data distribution. 

\begin{table}
  \centering
  % \begin{tabular}{@{}lc@{}}
  \resizebox{1\columnwidth}{!}{
  \begin{tabular}{cccccc}
    \toprule
    \multirow{2}*{Target motion}&\multirow{2}*{OneShot(\%)}& \multicolumn{3}{c}{FewShot(\%)} & \multirow{2}*{$\mathcal{M}_{\hat{train}}$(\%)}\\ \cline{3-5} 
        && 1\% & 5\% & 10\% &\\
    \midrule
    A0 & 62.84 & 84.09 & 88.34 & 91.62 & \multirow{6}*{91.80} \\ 
    A1 & 53.31& 79.54 & 89.74 & 90.04 &  \\  
    A2 & 57.62 &  78.14& 89.01 & 91.32 & \\ 
    A3 & 57.07& 81.30& 89.25 & 91.44 & \\
    A4 & 45.96& 82.33 & 88.16& 91.14 & \\
    A5 & 61.57 & 83.49 & 88.46 & 91.01 & \\
    \bottomrule
  \end{tabular}}
  \caption{Top-1 accuracy comparison. Sampling one-shot and few-shot (1\%, 5\%, and 10\%) from $\mathcal{M}_{\hat{train}}$ for action generation. The generated and sampled actions are concatenated into a new train set to train and test the ST-GCN. }
  \label{tab:2}
\end{table}

\subsection{Ablation Study and Comparison with Prior Work}

We conduct an ablation study of the loss term and a comparison with other methods to verify the validity of the loss term in the model and the state-of-the-art of our method. 
Specifically, we quantitatively measure the generation quality and Accuracy of the five generative models: 
$[$Aberman et al. 2020$]$, 
$[$Jang et al. 2022$]$, 
MGN($\mathcal{L}_{rec}+\mathcal{L}_{cyc}$), 
MGN($\mathcal{L}_{rec}+\mathcal{L}_{cyc}+\mathcal{L}_{trip}$), and AGN (Table. \ref{tab:3}).
Where MGN($\mathcal{L}_{rec}+\mathcal{L}_{cyc}+\mathcal{L}_{trip}$) is a part of AGN. Therefore, the FMD is both 2.67. 
Due to the effect of UMN, the recognition accuracy of AGN is 4.76\% higher than the former, which is 87.42\% and 92.18\%, respectively. 
The FMD and Accuracy are 2.90 and 83.43\% for the MGN without $\mathcal{L}_{trip}$, proving that our design of $\mathcal{L}_{trip}$ is effective in generative networks. The method of Aberman et al. measures the FMD to be 21.36 and the Accuracy to be 51.34\%. Motion Puzzle measured an FMD of 9.42 and an Accuracy of 67.63\%.
In comparison, our method is competitive.

\begin{table}
  \centering
  % \begin{tabular}{@{}lc@{}}
  \resizebox{1\columnwidth}{!}{
  \begin{tabular}{lcc}
    \toprule
    Methods&FMD$\downarrow$&Acc(\%)$\uparrow$ \\
    \midrule
    $[$Aberman et al. 2020$]$& 21.36 $\pm$ 2.37& 51.34 $\pm$ 1.92\\
    $[$Jang et al. 2022$]$& 9.42 $\pm$ 0.72& 67.63 $\pm$ 3.95 \\
    \midrule
    MGN ($\mathcal{L}_{rec}+\mathcal{L}_{cyc}$) & 2.90 $\pm$ 0.54& 83.43 $\pm$ 3.21 \\ 
    MGN ($\mathcal{L}_{rec}+\mathcal{L}_{cyc}+\mathcal{L}_{trip}$) & 2.67 $\pm$ 0.37& 87.42 $\pm$ 2.24 \\
    \textbf{AGN (Ours)}& \textbf{2.67 $\pm$ 0.37}& \textbf{92.18 $\pm$ 2.75} \\
    \bottomrule
  \end{tabular}}
  \caption{FMD and Acc are measured using five methods: [Aberman et al. 2020], [Jang et al. 2022], MGN($\mathcal{L}_{rec}+\mathcal{L}_{cyc}$), MGN($\mathcal{L}_{rec}+\mathcal{L}_{cyc}+\mathcal{L}_{trip}$), and AGN.}
  \label{tab:3}
\end{table}

\begin{figure*}[htpb]
  \centering
   \includegraphics[width=0.8\linewidth]{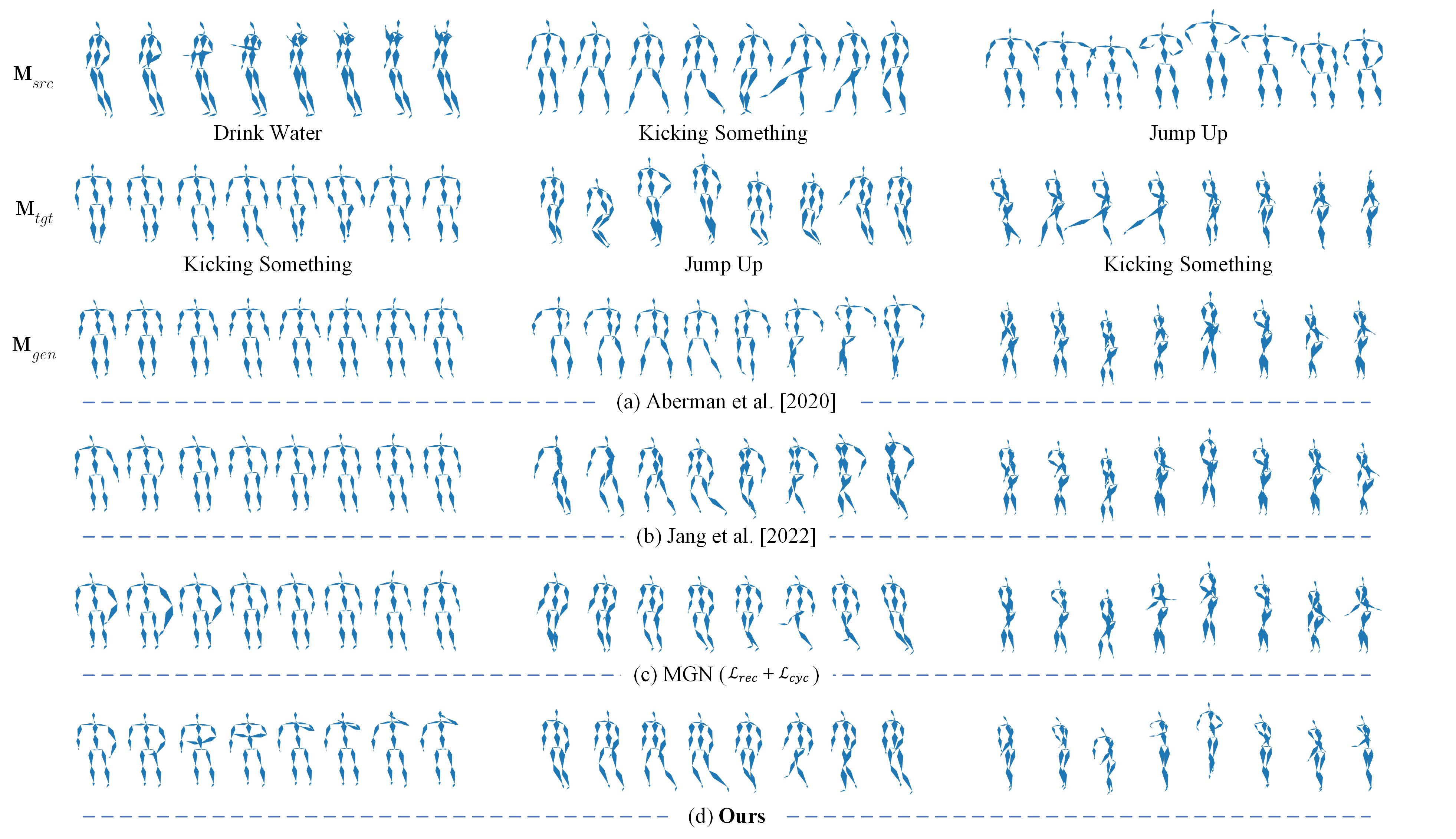}

   \caption{Comparison results. We used four methods to generate hand action (“Drinking Water”), leg action (“Kicking Something”), and whole-body action (“Jump up”).}
   \label{fig:6}
\end{figure*}

Figure. \ref{fig:6} shows the actions generated by the four methods: (a) the actions generated by Aberman's method, (b) the actions generated by Motion Puzzle, (c) the actions generated by MGN ($\mathcal{L}_{trip}$), and (d) the actions generated by our method.  
Compared to other methods, our method is optimal in generating hand action (“Drinking Water”), leg action (“Kicking Something”), and whole-body action (“Jump up”). 
In (a), (b), and (c), the “Drink Water” (first column) contains only the body morphology of the target action but not the hand moves of the source action. 
The “kicking Something” (second column) and “Jump up” (third column) combine the category features of the source action and the morphology features of the target action very well, but the hand moves are very raw.
Our method generates more natural and coordinated actions, both in terms of the moves of the parts and the overall morphology.

To fully demonstrate that UMN is effective, we sampled 1\% from $\mathcal{M}_{\hat{train}}$ and generated actions using MGN and MGN+UMN, respectively. Then, it is tested according to the action recognition experiment (sec. \ref{sec:motion recognition}). 
Table \ref{tab:4} shows that the average recognition accuracy is 81.48\% for MGN+UMN and 64.96\% for MGN. The results demonstrate that the UMN is effective.

\begin{table}
  \centering
  % \begin{tabular}{@{}lc@{}}
  \resizebox{1\columnwidth}{!}{
  \begin{tabular}{lcccccc}
    \toprule
    & A0 &  A1 & A2 & A3 & A4 & A5  \\
    \midrule
    MGN($\mathcal{L}_{total}$)& 58.29 & 62.72 & 74.62 & 57.32 & 67.88 & 68.91 \\
    MGN($\mathcal{L}_{total}$)+UMN& 84.09 & 79.54 & 78.14 & 81.30 & 82.33 & 83.49 \\
    \bottomrule
  \end{tabular}}
  \caption{Comparison of Top-1 accuracy of MGN($\mathcal{L}_{total}$) and MGN($\mathcal{L}_{total}$)+UMN. Sampling 1\% from $\mathcal{M}_{\hat{train}}$ for action generation.}
  \label{tab:4}
\end{table}

\section{Conclusion}

In this paper, we propose a novel generative network called AGN by introducing active learning. 
AGN can generate many new actions by means of motion transfer with only one or a few samples. 
The AGN consists of the MGN and the UMN. 
The MGN is able to implicitly learn the skeletal morphology of the target action while preserving the category features of the source action. 
The UMN utilizes uncertainty-inspired learning in active learning to provide an uncertainty score for the generation process and thus guides the MGN to improve the quality of the generation. 
AGN showed the best performance compared to the existing methods.FMD is 2.67, and Accuracy is 92.18\%. 
{
    \small
    \bibliographystyle{ieeenat_fullname}
    \bibliography{main}

\begin{thebibliography}{72}
\providecommand{\natexlab}[1]{#1}
\providecommand{\url}[1]{\texttt{#1}}
\expandafter\ifx\csname urlstyle\endcsname\relax
  \providecommand{\doi}[1]{doi: #1}\else
  \providecommand{\doi}{doi: \begingroup \urlstyle{rm}\Url}\fi

\bibitem[Aberman et~al.(2020)Aberman, Weng, Lischinski, Cohen-Or, and Chen]{aberman2020unpaired}
Kfir Aberman, Yijia Weng, Dani Lischinski, Daniel Cohen-Or, and Baoquan Chen.
\newblock Unpaired motion style transfer from video to animation.
\newblock \emph{ACM Transactions on Graphics (TOG)}, 39\penalty0 (4):\penalty0 64--1, 2020.

\bibitem[Aggarwal and Xia(2014)]{aggarwal2014human}
Jake~K Aggarwal and Lu Xia.
\newblock Human activity recognition from 3d data: A review.
\newblock \emph{Pattern Recognition Letters}, 48:\penalty0 70--80, 2014.

\bibitem[Antoniou et~al.(2017)Antoniou, Storkey, and Edwards]{antoniou2017data}
Antreas Antoniou, Amos Storkey, and Harrison Edwards.
\newblock Data augmentation generative adversarial networks.
\newblock \emph{arXiv preprint arXiv:1711.04340}, 2017.

\bibitem[Bai et~al.(2022)Bai, Ding, Xu, Chi, Zhang, and Sun]{bai2022skeleton}
Zhongyu Bai, Qichuan Ding, Hongli Xu, Jianning Chi, Xiangyue Zhang, and Tiansheng Sun.
\newblock Skeleton-based similar action recognition through integrating the salient image feature into a center-connected graph convolutional network.
\newblock \emph{Neurocomputing}, 507:\penalty0 40--53, 2022.

\bibitem[Beluch et~al.(2018)Beluch, Genewein, N{\"u}rnberger, and K{\"o}hler]{beluch2018power}
William~H Beluch, Tim Genewein, Andreas N{\"u}rnberger, and Jan~M K{\"o}hler.
\newblock The power of ensembles for active learning in image classification.
\newblock In \emph{Proceedings of the IEEE conference on computer vision and pattern recognition}, pages 9368--9377, 2018.

\bibitem[Cai et~al.(2021)Cai, Jiang, Han, Jia, and Lu]{cai2021jolo}
Jinmiao Cai, Nianjuan Jiang, Xiaoguang Han, Kui Jia, and Jiangbo Lu.
\newblock Jolo-gcn: mining joint-centered light-weight information for skeleton-based action recognition.
\newblock In \emph{Proceedings of the IEEE/CVF winter conference on applications of computer vision}, pages 2735--2744, 2021.

\bibitem[Carreira and Zisserman(2017)]{carreira2017quo}
Joao Carreira and Andrew Zisserman.
\newblock Quo vadis, action recognition? a new model and the kinetics dataset.
\newblock In \emph{proceedings of the IEEE Conference on Computer Vision and Pattern Recognition}, pages 6299--6308, 2017.

\bibitem[Choi et~al.(2018)Choi, Choi, Kim, Ha, Kim, and Choo]{choi2018stargan}
Yunjey Choi, Minje Choi, Munyoung Kim, Jung-Woo Ha, Sunghun Kim, and Jaegul Choo.
\newblock Stargan: Unified generative adversarial networks for multi-domain image-to-image translation.
\newblock In \emph{Proceedings of the IEEE conference on computer vision and pattern recognition}, pages 8789--8797, 2018.

\bibitem[Choi et~al.(2020)Choi, Uh, Yoo, and Ha]{choi2020stargan}
Yunjey Choi, Youngjung Uh, Jaejun Yoo, and Jung-Woo Ha.
\newblock Stargan v2: Diverse image synthesis for multiple domains.
\newblock In \emph{Proceedings of the IEEE/CVF conference on computer vision and pattern recognition}, pages 8188--8197, 2020.

\bibitem[Clou{\^a}tre and Demers(2019)]{clouatre2019figr}
Louis Clou{\^a}tre and Marc Demers.
\newblock Figr: Few-shot image generation with reptile.
\newblock \emph{arXiv preprint arXiv:1901.02199}, 2019.

\bibitem[Du et~al.(2019)Du, Herrmann, Sprenger, Fischer, and Slusallek]{du2019stylistic}
Han Du, Erik Herrmann, Janis Sprenger, Klaus Fischer, and Philipp Slusallek.
\newblock Stylistic locomotion modeling and synthesis using variational generative models.
\newblock In \emph{Proceedings of the 12th ACM SIGGRAPH Conference on Motion, Interaction and Games}, pages 1--10, 2019.

\bibitem[Du et~al.(2021)Du, Zhao, Chen, Chai, Chen, and Li]{du2021contrastive}
Pan Du, Suyun Zhao, Hui Chen, Shuwen Chai, Hong Chen, and Cuiping Li.
\newblock Contrastive coding for active learning under class distribution mismatch.
\newblock In \emph{Proceedings of the IEEE/CVF International Conference on Computer Vision}, pages 8927--8936, 2021.

\bibitem[Ebrahimi et~al.(2019)Ebrahimi, Elhoseiny, Darrell, and Rohrbach]{ebrahimi2019uncertainty}
Sayna Ebrahimi, Mohamed Elhoseiny, Trevor Darrell, and Marcus Rohrbach.
\newblock Uncertainty-guided continual learning with bayesian neural networks.
\newblock \emph{arXiv preprint arXiv:1906.02425}, 2019.

\bibitem[Escalera et~al.(2017)Escalera, Athitsos, and Guyon]{escalera2017challenges}
Sergio Escalera, Vassilis Athitsos, and Isabelle Guyon.
\newblock Challenges in multi-modal gesture recognition.
\newblock \emph{Gesture recognition}, pages 1--60, 2017.

\bibitem[Feichtenhofer et~al.(2016)Feichtenhofer, Pinz, and Zisserman]{feichtenhofer2016convolutional}
Christoph Feichtenhofer, Axel Pinz, and Andrew Zisserman.
\newblock Convolutional two-stream network fusion for video action recognition.
\newblock In \emph{Proceedings of the IEEE conference on computer vision and pattern recognition}, pages 1933--1941, 2016.

\bibitem[Feichtenhofer et~al.(2019)Feichtenhofer, Fan, Malik, and He]{feichtenhofer2019slowfast}
Christoph Feichtenhofer, Haoqi Fan, Jitendra Malik, and Kaiming He.
\newblock Slowfast networks for video recognition.
\newblock In \emph{Proceedings of the IEEE/CVF international conference on computer vision}, pages 6202--6211, 2019.

\bibitem[Finn et~al.(2017)Finn, Abbeel, and Levine]{finn2017model}
Chelsea Finn, Pieter Abbeel, and Sergey Levine.
\newblock Model-agnostic meta-learning for fast adaptation of deep networks.
\newblock In \emph{International conference on machine learning}, pages 1126--1135. PMLR, 2017.

\bibitem[Gal and Ghahramani(2016)]{gal2016dropout}
Yarin Gal and Zoubin Ghahramani.
\newblock Dropout as a bayesian approximation: Representing model uncertainty in deep learning.
\newblock In \emph{international conference on machine learning}, pages 1050--1059. PMLR, 2016.

\bibitem[Gal et~al.(2017)Gal, Islam, and Ghahramani]{gal2017deep}
Yarin Gal, Riashat Islam, and Zoubin Ghahramani.
\newblock Deep bayesian active learning with image data.
\newblock In \emph{International conference on machine learning}, pages 1183--1192. PMLR, 2017.

\bibitem[Gatys et~al.(2016)Gatys, Ecker, and Bethge]{gatys2016image}
Leon~A Gatys, Alexander~S Ecker, and Matthias Bethge.
\newblock Image style transfer using convolutional neural networks.
\newblock In \emph{Proceedings of the IEEE conference on computer vision and pattern recognition}, pages 2414--2423, 2016.

\bibitem[Gorriz et~al.(2017)Gorriz, Carlier, Faure, and Gir{\'o}-i Nieto]{gorriz2017cost}
Marc Gorriz, Axel Carlier, Emmanuel Faure, and Xavier Gir{\'o}-i Nieto.
\newblock Cost-effective active learning for melanoma segmentation.
\newblock \emph{arXiv preprint arXiv:1711.09168}, 2017.

\bibitem[Han et~al.(2017)Han, Reily, Hoff, and Zhang]{han2017space}
Fei Han, Brian Reily, William Hoff, and Hao Zhang.
\newblock Space-time representation of people based on 3d skeletal data: A review.
\newblock \emph{Computer Vision and Image Understanding}, 158:\penalty0 85--105, 2017.

\bibitem[He et~al.(2019)He, Rakin, and Fan]{he2019parametric}
Zhezhi He, Adnan~Siraj Rakin, and Deliang Fan.
\newblock Parametric noise injection: Trainable randomness to improve deep neural network robustness against adversarial attack.
\newblock In \emph{Proceedings of the IEEE/CVF Conference on Computer Vision and Pattern Recognition}, pages 588--597, 2019.

\bibitem[Ho et~al.(2020)Ho, Jain, and Abbeel]{ho2020denoising}
Jonathan Ho, Ajay Jain, and Pieter Abbeel.
\newblock Denoising diffusion probabilistic models.
\newblock \emph{Advances in neural information processing systems}, 33:\penalty0 6840--6851, 2020.

\bibitem[Holden et~al.(2016)Holden, Saito, and Komura]{2016A}
Daniel Holden, Jun Saito, and Taku Komura.
\newblock A deep learning framework for character motion synthesis and editing.
\newblock \emph{Acm Transactions on Graphics}, 35\penalty0 (4):\penalty0 1--11, 2016.

\bibitem[Hong et~al.(2020{\natexlab{a}})Hong, Niu, Zhang, and Zhang]{hong2020matchinggan}
Yan Hong, Li Niu, Jianfu Zhang, and Liqing Zhang.
\newblock Matchinggan: Matching-based few-shot image generation.
\newblock In \emph{2020 IEEE International conference on multimedia and expo (ICME)}, pages 1--6. IEEE, 2020{\natexlab{a}}.

\bibitem[Hong et~al.(2020{\natexlab{b}})Hong, Niu, Zhang, Zhao, Fu, and Zhang]{hong2020f2gan}
Yan Hong, Li Niu, Jianfu Zhang, Weijie Zhao, Chen Fu, and Liqing Zhang.
\newblock F2gan: Fusing-and-filling gan for few-shot image generation.
\newblock In \emph{Proceedings of the 28th ACM international conference on multimedia}, pages 2535--2543, 2020{\natexlab{b}}.

\bibitem[Hong et~al.(2022)Hong, Niu, Zhang, and Zhang]{hong2022deltagan}
Yan Hong, Li Niu, Jianfu Zhang, and Liqing Zhang.
\newblock Deltagan: Towards diverse few-shot image generation with sample-specific delta.
\newblock In \emph{European Conference on Computer Vision}, pages 259--276. Springer, 2022.

\bibitem[Huang and Belongie(2017)]{huang2017arbitrary}
Xun Huang and Serge Belongie.
\newblock Arbitrary style transfer in real-time with adaptive instance normalization.
\newblock In \emph{Proceedings of the IEEE international conference on computer vision}, pages 1501--1510, 2017.

\bibitem[Jang et~al.(2022)Jang, Park, and Lee]{jang2022motion}
Deok-Kyeong Jang, Soomin Park, and Sung-Hee Lee.
\newblock Motion puzzle: Arbitrary motion style transfer by body part.
\newblock \emph{ACM Transactions on Graphics (TOG)}, 41\penalty0 (3):\penalty0 1--16, 2022.

\bibitem[Ji et~al.(2012)Ji, Xu, Yang, and Yu]{ji20123d}
Shuiwang Ji, Wei Xu, Ming Yang, and Kai Yu.
\newblock 3d convolutional neural networks for human action recognition.
\newblock \emph{IEEE transactions on pattern analysis and machine intelligence}, 35\penalty0 (1):\penalty0 221--231, 2012.

\bibitem[Karras et~al.(2019)Karras, Laine, and Aila]{karras2019style}
Tero Karras, Samuli Laine, and Timo Aila.
\newblock A style-based generator architecture for generative adversarial networks.
\newblock In \emph{Proceedings of the IEEE/CVF conference on computer vision and pattern recognition}, pages 4401--4410, 2019.

\bibitem[Karras et~al.(2020)Karras, Laine, Aittala, Hellsten, Lehtinen, and Aila]{karras2020analyzing}
Tero Karras, Samuli Laine, Miika Aittala, Janne Hellsten, Jaakko Lehtinen, and Timo Aila.
\newblock Analyzing and improving the image quality of stylegan.
\newblock In \emph{Proceedings of the IEEE/CVF conference on computer vision and pattern recognition}, pages 8110--8119, 2020.

\bibitem[Kingma and Dhariwal(2018)]{kingma2018glow}
Durk~P Kingma and Prafulla Dhariwal.
\newblock Glow: Generative flow with invertible 1x1 convolutions.
\newblock \emph{Advances in neural information processing systems}, 31, 2018.

\bibitem[Kingma and Welling(2013)]{kingma2013auto}
Diederik~P Kingma and Max Welling.
\newblock Auto-encoding variational bayes.
\newblock \emph{arXiv preprint arXiv:1312.6114}, 2013.

\bibitem[Li et~al.(2019)Li, Chen, Chen, Zhang, Wang, and Tian]{li2019actional}
Maosen Li, Siheng Chen, Xu Chen, Ya Zhang, Yanfeng Wang, and Qi Tian.
\newblock Actional-structural graph convolutional networks for skeleton-based action recognition.
\newblock In \emph{Proceedings of the IEEE/CVF conference on computer vision and pattern recognition}, pages 3595--3603, 2019.

\bibitem[Liu et~al.(2019)Liu, Gu, and Samaras]{liu2019wasserstein}
Huidong Liu, Xianfeng Gu, and Dimitris Samaras.
\newblock Wasserstein gan with quadratic transport cost.
\newblock In \emph{Proceedings of the IEEE/CVF international conference on computer vision}, pages 4832--4841, 2019.

\bibitem[Liu et~al.(2022)Liu, Li, Du, Torralba, and Tenenbaum]{liu2022compositional}
Nan Liu, Shuang Li, Yilun Du, Antonio Torralba, and Joshua~B Tenenbaum.
\newblock Compositional visual generation with composable diffusion models.
\newblock In \emph{European Conference on Computer Vision}, pages 423--439. Springer, 2022.

\bibitem[Lopez et~al.(2018)Lopez, Regier, Jordan, and Yosef]{lopez2018information}
Romain Lopez, Jeffrey Regier, Michael~I Jordan, and Nir Yosef.
\newblock Information constraints on auto-encoding variational bayes.
\newblock \emph{Advances in neural information processing systems}, 31, 2018.

\bibitem[Lopez et~al.(2020)Lopez, Boyeau, Yosef, Jordan, and Regier]{lopez2020decision}
Romain Lopez, Pierre Boyeau, Nir Yosef, Michael Jordan, and Jeffrey Regier.
\newblock Decision-making with auto-encoding variational bayes.
\newblock \emph{Advances in Neural Information Processing Systems}, 33:\penalty0 5081--5092, 2020.

\bibitem[Mahapatra et~al.(2018)Mahapatra, Bozorgtabar, Thiran, and Reyes]{mahapatra2018efficient}
Dwarikanath Mahapatra, Behzad Bozorgtabar, Jean-Philippe Thiran, and Mauricio Reyes.
\newblock Efficient active learning for image classification and segmentation using a sample selection and conditional generative adversarial network.
\newblock In \emph{International Conference on Medical Image Computing and Computer-Assisted Intervention}, pages 580--588. Springer, 2018.

\bibitem[Mao et~al.(2017)Mao, Li, Xie, Lau, Wang, and Paul~Smolley]{mao2017least}
Xudong Mao, Qing Li, Haoran Xie, Raymond~YK Lau, Zhen Wang, and Stephen Paul~Smolley.
\newblock Least squares generative adversarial networks.
\newblock In \emph{Proceedings of the IEEE international conference on computer vision}, pages 2794--2802, 2017.

\bibitem[Mayer and Timofte(2020)]{mayer2020adversarial}
Christoph Mayer and Radu Timofte.
\newblock Adversarial sampling for active learning.
\newblock In \emph{Proceedings of the IEEE/CVF Winter Conference on Applications of Computer Vision}, pages 3071--3079, 2020.

\bibitem[Mogren(2016)]{mogren2016c}
Olof Mogren.
\newblock C-rnn-gan: Continuous recurrent neural networks with adversarial training.
\newblock \emph{arXiv preprint arXiv:1611.09904}, 2016.

\bibitem[Nguyen and Smeulders(2004)]{nguyen2004active}
Hieu~T Nguyen and Arnold Smeulders.
\newblock Active learning using pre-clustering.
\newblock In \emph{Proceedings of the twenty-first international conference on Machine learning}, page~79, 2004.

\bibitem[Nichol and Schulman(2018)]{nichol2018reptile}
Alex Nichol and John Schulman.
\newblock Reptile: a scalable metalearning algorithm.
\newblock \emph{arXiv preprint arXiv:1803.02999}, 2\penalty0 (3):\penalty0 4, 2018.

\bibitem[Pan et~al.(2023)Pan, Tewari, Leimk{\"u}hler, Liu, Meka, and Theobalt]{pan2023drag}
Xingang Pan, Ayush Tewari, Thomas Leimk{\"u}hler, Lingjie Liu, Abhimitra Meka, and Christian Theobalt.
\newblock Drag your gan: Interactive point-based manipulation on the generative image manifold.
\newblock In \emph{ACM SIGGRAPH 2023 Conference Proceedings}, pages 1--11, 2023.

\bibitem[ParkSoomin et~al.(2021)ParkSoomin, JangDeok-Kyeong, and LeeSung-Hee]{ParkSoomin2021Diverse}
ParkSoomin, JangDeok-Kyeong, and LeeSung-Hee.
\newblock Diverse motion stylization for multiple style domains via spatial-temporal graph-based generative model.
\newblock \emph{Proceedings of the ACM on Computer Graphics and Interactive Techniques}, 2021.

\bibitem[Patashnik et~al.(2021)Patashnik, Wu, Shechtman, Cohen-Or, and Lischinski]{patashnik2021styleclip}
Or Patashnik, Zongze Wu, Eli Shechtman, Daniel Cohen-Or, and Dani Lischinski.
\newblock Styleclip: Text-driven manipulation of stylegan imagery.
\newblock In \emph{Proceedings of the IEEE/CVF International Conference on Computer Vision}, pages 2085--2094, 2021.

\bibitem[Radford et~al.(2015)Radford, Metz, and Chintala]{radford2015unsupervised}
Alec Radford, Luke Metz, and Soumith Chintala.
\newblock Unsupervised representation learning with deep convolutional generative adversarial networks.
\newblock \emph{arXiv preprint arXiv:1511.06434}, 2015.

\bibitem[Rezende and Mohamed(2015)]{rezende2015variational}
Danilo Rezende and Shakir Mohamed.
\newblock Variational inference with normalizing flows.
\newblock In \emph{International conference on machine learning}, pages 1530--1538. PMLR, 2015.

\bibitem[Saito et~al.(2020)Saito, Saenko, and Liu]{saito2020coco}
Kuniaki Saito, Kate Saenko, and Ming-Yu Liu.
\newblock Coco-funit: Few-shot unsupervised image translation with a content conditioned style encoder.
\newblock In \emph{Computer Vision--ECCV 2020: 16th European Conference, Glasgow, UK, August 23--28, 2020, Proceedings, Part III 16}, pages 382--398. Springer, 2020.

\bibitem[Saito et~al.(2017)Saito, Matsumoto, and Saito]{saito2017temporal}
Masaki Saito, Eiichi Matsumoto, and Shunta Saito.
\newblock Temporal generative adversarial nets with singular value clipping.
\newblock In \emph{Proceedings of the IEEE international conference on computer vision}, pages 2830--2839, 2017.

\bibitem[Shahroudy et~al.(2016)Shahroudy, Liu, Ng, and Wang]{shahroudy2016ntu}
Amir Shahroudy, Jun Liu, Tian-Tsong Ng, and Gang Wang.
\newblock Ntu rgb+ d: A large scale dataset for 3d human activity analysis.
\newblock In \emph{Proceedings of the IEEE conference on computer vision and pattern recognition}, pages 1010--1019, 2016.

\bibitem[Shao and Zhang(2021)]{shao2021spatchgan}
Xuning Shao and Weidong Zhang.
\newblock Spatchgan: A statistical feature based discriminator for unsupervised image-to-image translation.
\newblock In \emph{Proceedings of the IEEE/CVF International Conference on Computer Vision}, pages 6546--6555, 2021.

\bibitem[Shi et~al.(2019)Shi, Zhang, Cheng, and Lu]{shi2019two}
Lei Shi, Yifan Zhang, Jian Cheng, and Hanqing Lu.
\newblock Two-stream adaptive graph convolutional networks for skeleton-based action recognition.
\newblock In \emph{Proceedings of the IEEE/CVF conference on computer vision and pattern recognition}, pages 12026--12035, 2019.

\bibitem[Shorten and Khoshgoftaar(2019)]{shorten2019survey}
Connor Shorten and Taghi~M Khoshgoftaar.
\newblock A survey on image data augmentation for deep learning.
\newblock \emph{Journal of big data}, 6\penalty0 (1):\penalty0 1--48, 2019.

\bibitem[Simonyan and Zisserman(2014)]{simonyan2014very}
Karen Simonyan and Andrew Zisserman.
\newblock Very deep convolutional networks for large-scale image recognition.
\newblock \emph{arXiv preprint arXiv:1409.1556}, 2014.

\bibitem[Sinha et~al.(2019)Sinha, Ebrahimi, and Darrell]{sinha2019variational}
Samarth Sinha, Sayna Ebrahimi, and Trevor Darrell.
\newblock Variational adversarial active learning.
\newblock In \emph{Proceedings of the IEEE/CVF International Conference on Computer Vision}, pages 5972--5981, 2019.

\bibitem[Vinker et~al.(2021)Vinker, Horwitz, Zabari, and Hoshen]{vinker2021image}
Yael Vinker, Eliahu Horwitz, Nir Zabari, and Yedid Hoshen.
\newblock Image shape manipulation from a single augmented training sample.
\newblock In \emph{Proceedings of the IEEE/CVF International Conference on Computer Vision}, pages 13769--13778, 2021.

\bibitem[Vondrick et~al.(2016)Vondrick, Pirsiavash, and Torralba]{vondrick2016generating}
Carl Vondrick, Hamed Pirsiavash, and Antonio Torralba.
\newblock Generating videos with scene dynamics.
\newblock \emph{Advances in neural information processing systems}, 29, 2016.

\bibitem[Wang et~al.(2018{\natexlab{a}})Wang, Zhou, Tang, Fu, Tian, and Li]{wang2018unregularized}
Jiayu Wang, Wengang Zhou, Jinhui Tang, Zhongqian Fu, Qi Tian, and Houqiang Li.
\newblock Unregularized auto-encoder with generative adversarial networks for image generation.
\newblock In \emph{Proceedings of the 26th ACM international conference on Multimedia}, pages 709--717, 2018{\natexlab{a}}.

\bibitem[Wang et~al.(2018{\natexlab{b}})Wang, Li, Ogunbona, Wan, and Escalera]{wang2018rgb}
Pichao Wang, Wanqing Li, Philip Ogunbona, Jun Wan, and Sergio Escalera.
\newblock Rgb-d-based human motion recognition with deep learning: A survey.
\newblock \emph{Computer vision and image understanding}, 171:\penalty0 118--139, 2018{\natexlab{b}}.

\bibitem[Xia et~al.(2015)Xia, Wang, Chai, and Hodgins]{xia2015realtime}
Shihong Xia, Congyi Wang, Jinxiang Chai, and Jessica Hodgins.
\newblock Realtime style transfer for unlabeled heterogeneous human motion.
\newblock \emph{ACM Transactions on Graphics (TOG)}, 34\penalty0 (4):\penalty0 1--10, 2015.

\bibitem[Yan et~al.(2018)Yan, Xiong, and Lin]{yan2018spatial}
Sijie Yan, Yuanjun Xiong, and Dahua Lin.
\newblock Spatial temporal graph convolutional networks for skeleton-based action recognition.
\newblock In \emph{Proceedings of the AAAI conference on artificial intelligence}, 2018.

\bibitem[Yan et~al.(2019)Yan, Li, Xiong, Yan, and Lin]{yan2019convolutional}
Sijie Yan, Zhizhong Li, Yuanjun Xiong, Huahan Yan, and Dahua Lin.
\newblock Convolutional sequence generation for skeleton-based action synthesis.
\newblock In \emph{Proceedings of the IEEE/CVF International Conference on Computer Vision}, pages 4394--4402, 2019.

\bibitem[Yang et~al.(2019)Yang, Yuan, Li, Du, Xing, Hu, and Maybank]{yang2019asymmetric}
Hao Yang, Chunfeng Yuan, Bing Li, Yang Du, Junliang Xing, Weiming Hu, and Stephen~J Maybank.
\newblock Asymmetric 3d convolutional neural networks for action recognition.
\newblock \emph{Pattern recognition}, 85:\penalty0 1--12, 2019.

\bibitem[Yang et~al.(2017)Yang, Zhang, Chen, Zhang, and Chen]{yang2017suggestive}
Lin Yang, Yizhe Zhang, Jianxu Chen, Siyuan Zhang, and Danny~Z Chen.
\newblock Suggestive annotation: A deep active learning framework for biomedical image segmentation.
\newblock In \emph{Medical Image Computing and Computer Assisted Intervention- MICCAI 2017: 20th International Conference, Quebec City, QC, Canada, September 11-13, 2017, Proceedings, Part III 20}, pages 399--407. Springer, 2017.

\bibitem[Yumer and Mitra(2016)]{yumer2016spectral}
M~Ersin Yumer and Niloy~J Mitra.
\newblock Spectral style transfer for human motion between independent actions.
\newblock \emph{ACM Transactions on Graphics (TOG)}, 35\penalty0 (4):\penalty0 1--8, 2016.

\bibitem[Zhang et~al.(2020)Zhang, Li, Yang, Wang, Zha, and Huang]{zhang2020state}
Beichen Zhang, Liang Li, Shijie Yang, Shuhui Wang, Zheng-Jun Zha, and Qingming Huang.
\newblock State-relabeling adversarial active learning.
\newblock In \emph{Proceedings of the IEEE/CVF conference on computer vision and pattern recognition}, pages 8756--8765, 2020.

\bibitem[Zhang et~al.(2023)Zhang, Huang, Tang, Huang, Ma, Dong, and Xu]{zhang2023inversion}
Yuxin Zhang, Nisha Huang, Fan Tang, Haibin Huang, Chongyang Ma, Weiming Dong, and Changsheng Xu.
\newblock Inversion-based style transfer with diffusion models.
\newblock In \emph{Proceedings of the IEEE/CVF Conference on Computer Vision and Pattern Recognition}, pages 10146--10156, 2023.

\bibitem[Zhu and Bento(2017)]{zhu2017generative}
Jia-Jie Zhu and Jos{\'e} Bento.
\newblock Generative adversarial active learning.
\newblock \emph{arXiv preprint arXiv:1702.07956}, 2017.

\end{thebibliography}
}

\end{document}